\documentclass{article}

\usepackage{microtype}
\usepackage{graphicx}
\usepackage{subfigure}
\usepackage{booktabs} 

\usepackage{hyperref}

\usepackage[accepted]{icml2024}

\usepackage{amsmath}
\usepackage{amssymb}
\usepackage{mathtools}
\usepackage{amsthm}
\usepackage{bbm}

\usepackage[capitalize,noabbrev]{cleveref}

\theoremstyle{plain}

\theoremstyle{definition}

\theoremstyle{remark}

\usepackage[textsize=tiny]{todonotes}

\usepackage{siunitx}
\usepackage[flushleft]{threeparttable}
\usepackage{tabularx}

\newcommand{\feat}[1]{\sc #1\normalfont}

\usepackage{tikz}
\usepackage[utf8]{inputenc}
\usepackage{pgfplots}
\DeclareUnicodeCharacter{2212}{−}
\usepgfplotslibrary{groupplots,dateplot}
\usetikzlibrary{patterns,shapes.arrows}
\pgfplotsset{compat=newest}

\icmltitlerunning{Graph Neural Networks and Spatial Information Learning for Post-Processing Ensemble Weather Forecasts}

\begin{document}

\twocolumn[
\icmltitle{\vspace{-1pt}Graph Neural Networks and Spatial Information Learning for\\ Post-Processing Ensemble Weather Forecasts\vspace{-1pt}}



\icmlsetsymbol{equal}{*}

\begin{icmlauthorlist}
\icmlauthor{Moritz Feik}{HITS,KIT}
\icmlauthor{Sebastian Lerch}{HITS,KIT,equal}
\icmlauthor{Jan Stühmer}{HITS,KIT,equal}
\end{icmlauthorlist}

\icmlaffiliation{KIT}{Karlsruhe Institute of Technology, Karlsruhe, Germany}
\icmlaffiliation{HITS}{Heidelberg Institute for Theoretical Studies, Heidelberg, Germany}

\icmlcorrespondingauthor{Moritz Feik}{moritz.feik@student.kit.edu}

\icmlkeywords{Machine Learning, ICML}

\vskip 0.3in
]



\printAffiliationsAndNotice{\icmlEqualContribution} 

\begin{abstract}
	Ensemble forecasts from numerical weather prediction models show systematic errors that require correction via post-processing.
	While there has been substantial progress in flexible neural network-based post-processing methods over the past years, most station-based approaches still treat every input data point separately which limits the capabilities for leveraging spatial structures in the forecast errors.
	In order to improve information sharing across locations, we propose a graph neural network architecture for ensemble post-processing, which represents the station locations as nodes on a graph and utilizes an attention mechanism to identify relevant predictive information from neighboring locations.
	In a case study on 2-m temperature forecasts over Europe, the graph neural network model shows substantial improvements over a highly competitive neural network-based post-processing method.
\end{abstract}

\section{Introduction}
\label{sec:introduction}

Modern weather forecasts utilize ensemble simulations from numerical weather prediction (NWP) models with different initial conditions or model physics.
Even though NWP ensemble predictions have seen substantial progress over the past decades \citep{BauerEtAl2015}, they often show systematic biases and fail to correctly quantify forecast uncertainty.
Therefore, statistical or machine learning methods are required to correct these errors in a process referred to as post-processing, which has become a standard practice in research and operations. 
Most modern post-processing methods yield forecast distributions as their output, e.g.\ in the form of parameters of a pre-specified family of probability distributions.
A major focus of post-processing research over the past years has been on flexible machine learning (ML) techniques which have demonstrated superior forecast performance, primarily due to their ability to incorporate additional predictor variables beyond ensemble forecasts of the target variable \citep{HauptEtAl2021,VannitsemEtAl2021}.
Specifically, neural network (NN)-based distributional regression approaches first proposed by \citet{rasp_neural_2018} have shown considerable success. Thereby, NNs enable the data-driven learning of nonlinear relationships between arbitrary predictor variables and forecast distribution parameters.
Over the past years, NN-based post-processing methods have been extended in several directions, including non-parametric approaches \citep{bremnes2020ensemble}, CNN-based methods for two-dimensional gridded forecast fields \citep{scheuerer_using_2020, veldkamp_statistical_2021,ChapmanEtAl2022,HoratLerch2024}, generative ML methods for multivariate post-processing \citep{ChenEtAl2024}, or permutation-invariant set transformer architectures to model interactions between individual ensemble members \citep{hohlein_postprocessing_2024}. 

The aforementioned CNN models incorporate spatial information between locations for gridded domains. Most station-based post-processing methods still treat every input data point separately, which prevents the models from sharing information across locations and thereby leveraging spatial structures in the forecast errors.
To address this limitation, we propose graph neural network (GNN) architectures for post-processing, where weather stations form the nodes on a graph.
By obtaining forecast distribution parameters in a node-level prediction setting, GNN-based post-processing methods are able to leverage spatial dependencies between stations and enable improved sharing of information across locations during model training and inference compared to standard NN approaches.

\section{Data}
\label{sec:data}

In order to facilitate a fair and standardized comparison to other methods, we use EUPPBench, a benchmark dataset for ensemble post-processing \cite{demaeyer_euppbench_2023}.
The dataset includes medium-range ensemble forecasts from the European Centre for Medium-Range Weather Forecasts (ECMWF) along with corresponding station observations over an extended period for multiple lead times. 
In total, the data spans from 1997 to 2018 and includes 122 weather stations in Europe, see \cref{fig:EUPP-Stations} for details. 
Motivated by typical development practices for post-processing methods in operational weather prediction at meteorological services, the dataset contains both reforecasts and forecasts. 
Reforecasts are NWP model runs for past dates, which are conducted to obtain a large archive of past forecasts for analyzing various properties of the NWP system. 
The EUPPBench dataset contains 4180 reforecasts with a reduced number of 11 ensemble members from 1997 to 2017. 
In addition, the EUPPBench dataset includes of 730 daily operational forecasts from 2017--2018, which consist of 51 ensemble members. 
For both parts, a total of 31 predictor variables is available. 
We refer to \citet{demaeyer_euppbench_2023} for details.

We here focus on forecasts of 2-meter temperature (\sc t2m\normalfont) and report results for lead times of \SIlist{24;72;120}{\hour} in the interest of brevity.
Given the structure of the EUPPBench dataset and following \citet{hohlein_postprocessing_2024}, we consider two setups for post-processing tasks: ``reforecast to reforecast'' (\textit{R2R}) and ``reforecast to forecast'' (\textit{R2F}). 
The \textit{R2R} task consists of fitting a post-processing models to the reforecast data from 1997--2013, and testing this model on reforecasts from 2014--2017, whereas the \textit{R2F} task aims applying the fitted model to the forecast data from 2017--2018. 
The \textit{R2F} task can be viewed as a typical pathway for developing a post-processing model in operational weather prediction, and comes with additional technical challenges, e.g., the need to account for varying numbers of ensemble members in the training and test data.
\cref{tab:training data} lists the sizes of the training, validation and test datasets.

\section{Methods}
\label{sec:methods}

\subsection{Forecast Evaluation}
\label{sec:eval}

The main evaluation metric in the post-processing literature is the continuous ranked probability score (CRPS) given by
$
\text{CRPS}(F,y) = \int_{-\infty}^{\infty} \left( F(z) - \mathbbm{1}(y \leq z)\right)^2 \text{d}z,
$
where $F$ is the cumulative distribution function of the forecast distribution, $y$ is the realizing observation, and $\mathbbm{1}$ denotes the indicator function \citep[e.g.,][]{gneiting_probabilistic_2014}.
The CRPS simultaneously evaluates calibration and sharpness of the forecast distribution, and can be computed in analytical form for ensembles and many parametric families \citep{JordanEtAl2019}.
To assess the statistical significance of score differences, we use tests of equal predictive performance \citep{DieboldMariano1995}.

\subsection{DRN}
\label{sec:drn}

We utilize the distributional regression network (DRN) model originally proposed in \citet{rasp_neural_2018} as a state-of-the-art benchmark for station-based post-processing, which remains widely used and yields highly competitive benchmark forecasts \citep{VannitsemEtAl2021,schulz_machine_2022,hohlein_postprocessing_2024}.
The DRN model essentially is a standard fully-connected feed-forward NN which outputs the parameters of a predictive distribution, in our case the location $\mu$ and scale $\sigma$ of a Gaussian distribution which has been demonstrated to be an appropriate choice for \feat{t2m} prediction. 
Summary statistics from the NWP ensemble predictions of various meteorological variables serve as inputs to the NN.
We estimate a single model jointly for all stations by optimizing the CRPS as a loss function.
Thereby, station embeddings which map the station identifiers to a vector of latent features are used as additional inputs to generate local adaptivity. 
Our specific implementation of DRN follows \citet{hohlein_postprocessing_2024}, see their Section 3 for details.

\subsection{GNN}
\label{sec:gnn}
\begin{figure*}[!hbt]
	\begin{center}
		\centerline{\includegraphics[width=0.99\textwidth]{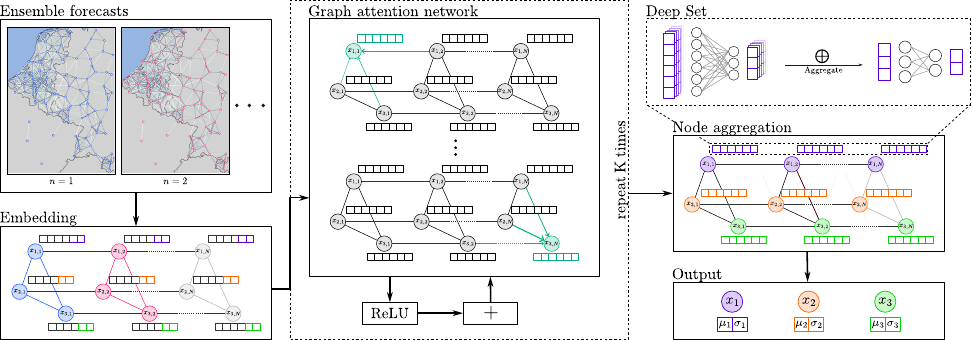}}
		\caption{Schematic illustration of the GNN model for ensemble post-processing. The input graph $\mathcal{G}$ is created from the $N$-member ensemble forecasts at $S$ stations. Next, the embedded station IDs are concatenated and passed to the GNN. The GNN block is repeated $K$ times with residual connections, followed by the node aggregation. Finally, a softplus function is applied to $\sigma$ to ensure positivity.}
		\label{fig:flow}
	\end{center}
	\vspace{-0.35in}
\end{figure*}

Graph neural networks (GNNs) are specialized deep learning models for graph-structured data, recognizing the value of representing problems in graph form rather than fixed grids or sequences \citep{gori2005new,scarselli2008graph}. Unlike traditional architectures, GNNs enable the modelling of complex interactions between nodes and edges within the graph.
\Cref{fig:flow} provides an overview of the proposed GNN model architecture. 
In a first step, the graph $\mathcal{G}$ is created, and, for each node, the  station identifier is replaced with its embedding, akin to the station embeddings in the DRN approach. 
The graph is then passed to $K$ GNN-blocks, which iteratively refine the hidden representations $\mathbf{h}_{s,n}$, where $s$ denotes the station and $n$ the member of the NWP ensemble.
Using $K$ blocks, each node incorporates information from $K$ hops away.
Each of these blocks has skip connections inspired by the ResNet model \cite{he_deep_2015}.
The residual learning approach helps to combat learning instabilities and leverage information from nodes multiple hops away.
After the hidden features are created, they are aggregated using the \textit{Deep Set} aggregation scheme \citep{zaheer_deep_2017}. 
For each station, the hidden features of the different ensemble members are used to compute the final outputs $\mu_s$ and $\sigma_s$.
The weights of all components of the GNN model are optimized jointly using the CRPS as a loss function. 

\subsubsection{Graph Topology}
\label{sec:graph topology}
In order for a GNN to process data, the data must be transformed into a graph.
For our dataset at hand, a graph $\mathcal{G}_t$ is created for each day $t$ for which a forecast exists. Each node $v_{s,n}$ represents the forecast for a particular station $s$ made by ensemble member $n$.
Additionally, each node carries the forecasts of several meteorological variables generated by the respective ensemble member as attributes.
Details are provided in \cref{tab:axuiliary variables,tab:station specific}. 
Stations that are closer than a certain threshold $d_{\text{max}}$ and stations with the same identifier are bidirectionally connected.
Each edge carries the normalized distance as a feature, while the edges between the ensemble members have a very small value $\epsilon$ instead of 0 as an attribute to facilitate training. Accordingly, the set of edges is
$
\mathcal{E}=\{(v_{i,u}, v_{j,v})\,\vert\, i=j \vee d(v_{i,u},v_{j,v})<d_{\text{max}}\}\,,
$
where $d(\cdot)$ is the geodesic distance.

\subsubsection{Graph Neural Networks}
\label{sec:GNN}

GNNs operate on the principle that they can learn and reason about graph-structured data by aggregating information from neighboring nodes and edges iteratively through message passing.
One of the many types of GNNs is the graph attentional network (GAT), which weights incoming messages for each node using an attention function $a$ \citep{velickovic_graph_2018, brody_how_2022}.
The hidden representations in GATs are generally computed as
$
\mathbf{h}_i = \phi \left(\mathbf{x}_i, \bigoplus_{j\in\ \mathcal{N}(i)} a(\mathbf{x}_i, \mathbf{x}_j)\psi(\mathbf{x}_j) \right)
$
\cite{bronstein_geometric_2021}.
For details on $\phi$, $\psi$, and $a$, see \citet{brody_how_2022}.
With the attention mechanism, each node is able to discern important from unimportant neighbors and aggregate only relevant messages.
Our implementation uses a GAT with \textit{multi-head attention} to stabilize learning, employing multiple independent attention mechanisms and concatenating their outputs for the new node representation \cite{vaswani_attention_2023}.\looseness=-1

\subsubsection{Permutation Invariant Node Aggregation}
\label{sec:deep set}
After processing the input graph $\mathcal{G}$, we generate predictions for $\mu_s$ and $\sigma_s$ based on the output of the GNN, which consists of the hidden features $\mathbf{h}_{s,n}, n = 1,...,N$.
Since the ensemble members are interchangeable, the aggregation along the ensemble dimension $n$ should be permutation invariant.
Such an aggregation scheme can be achieved by using \textit{Deep Sets} \cite{zaheer_deep_2017}.
Specifically, each set of hidden features for a given station $\mathcal{H}_s=\{\mathbf{h}_{s,n}, n=1,\dots, N\}$ is aggregated using 
$
(\mu_s,\sigma_s)=\rho\left(\frac{1}{N}\sum_{n=1}^{N}\phi(\mathbf{h}_{s,n})\right)
$
as an aggregation function. In our concrete implementation, $\rho$ and $\phi$ are both two-layer NNs.

\section{Results}
\label{sec:results}

\begin{table*}[htb]
	\caption{Scores for the reforecast to forecast task calculated per lead time, with the best CRPS scores highlighted in \textbf{bold}. The nominal level of the central prediction interval (PI) is $N-1/N+1$, where $N$ is the number of ensemble members.
		The coverage (PI COVER) is the ratio of how often the observation is contained in the PI and should be close to the nominal level for a calibrated forecast.
	}
	\label{tab:crps results}
	\vskip 0.1in
	\begin{center}
		\begin{small}
			\begin{sc}
				\begin{tabular}{lccccccccc}
					\toprule
					\textbf{Lead time} & \multicolumn{3}{c}{\SI{24}{\hour}} & \multicolumn{3}{c}{\SI{72}{\hour}} & \multicolumn{3}{c}{\SI{120}{\hour}}\\
					
					\cmidrule(lr){1-1}
					\cmidrule(lr){2-4}
					\cmidrule(lr){5-7}
					\cmidrule(lr){8-10}
					
					\textbf{Method} & CRPS & PI length & PI cover & CRPS & PI length & PI cover & CRPS & PI length & PI cover\\
					
					\midrule
					ENS & 1.12 & 2.66 & 56.06 & 1.18 & 4.72 & 72.90 & 1.38 & 7.14 & 81.16 \\
					DRN & 0.61 & 4.26 & 94.87 & 0.79 & 5.90 & 96.37 & 1.11 & 7.99 & 95.82\\[0.5em]
					SMRY & 0.62 & 4.45 & 95.53 & 0.79 & 6.17 & 97.01 & 1.10 & 8.31 & 96.64 \\
					DS & 0.61 & 4.41 & 95.72 & \textbf{0.78} & 4.43 & 89.87 & 1.14 & 4.56 & 77.79\\
					GAT & \textbf{0.60} & 4.16 & 95.04 & \textbf{0.78} & 5.93 & 96.42 & \textbf{1.09} & 8.27 & 96.80\\
					\bottomrule
				\end{tabular}
			\end{sc}
		\end{small}
	\end{center}
	\vspace{-0.25in}
\end{table*}

For our experiments we implemented the proposed method with the PyTorch Geometric~\cite{fey2019fastgraphrepresentationlearning} framework\footnote{The implementation can be downloaded from \url{https://github.com/hits-mli/gnn-post-processing}.}.
We evaluate the performance of the proposed model by training it on the EUPPBench dataset described in Section \ref{sec:data}. 
Here, we focus on the ``reforecast to forecast''~(\textit{R2F}) task. Additional, qualitatively similar results for the ``reforecast to reforecast''~(\textit{R2R}) task are available in the supplemental material.
Although the number of ensemble members, $N$, is arbitrary, we process the 51 ensemble members of the forecast data batch-wise in groups of 4 $\times$ 10 and a remaining group of 11, and average the predictions because the reforecasts used as training data contain only 11 ensemble members.
This procedure aims to better account for the different number of ensemble members in the reforecast and forecast data, and results in better forecast performance.
\Cref{tab:crps results} provides an overview of the results for the \textit{R2F} task. Results for the \textit{R2R} task are available in the supplementary material in \cref{tab:scores_full}.
We compare the proposed model (GAT) against a GNN model which only operates on one graph based on the summary statistics (i.e., mean and standard deviation) of the ensemble forecasts (SMRY), a pure Deep Set architecture (DS), where all edges from the initial graph except for self loops are removed, a fully-connected, feed-forward DRN model described in Section \ref{sec:drn}, and the unprocessed ensemble forecasts (ENS). For each lead time, we train a separate model.
These comparisons enable us to assess whether there is important information in the distribution of the NWP ensemble members and if the information sharing among weather station enabled by the GNN improves performance.

Not surprisingly, all post-processing methods substantially improve the raw ensemble predictions, which provide the sharpest prediction intervals, but fail to achieve a coverage close to the nominal value and thus clearly lack calibration. 
The proposed GAT model outperforms all other post-processing models in terms of the mean CRPS across all lead times and tasks. 
The statistical significance of these improvements is assessed via formal statistical tests following \citet{DieboldMariano1995}. Detailed results available in the supplemental material indicate that the improvements achieved by the GAT model are significant at the 5\% level for a large fraction of the investigated stations and lead times.
Interestingly, the DS model produces substantially sharper prediction intervals at longer lead times, but fails to achieve improvements over the DRN model in terms of the CRPS.

\begin{figure}
	\begin{center}
		\includegraphics[width=0.9\columnwidth]{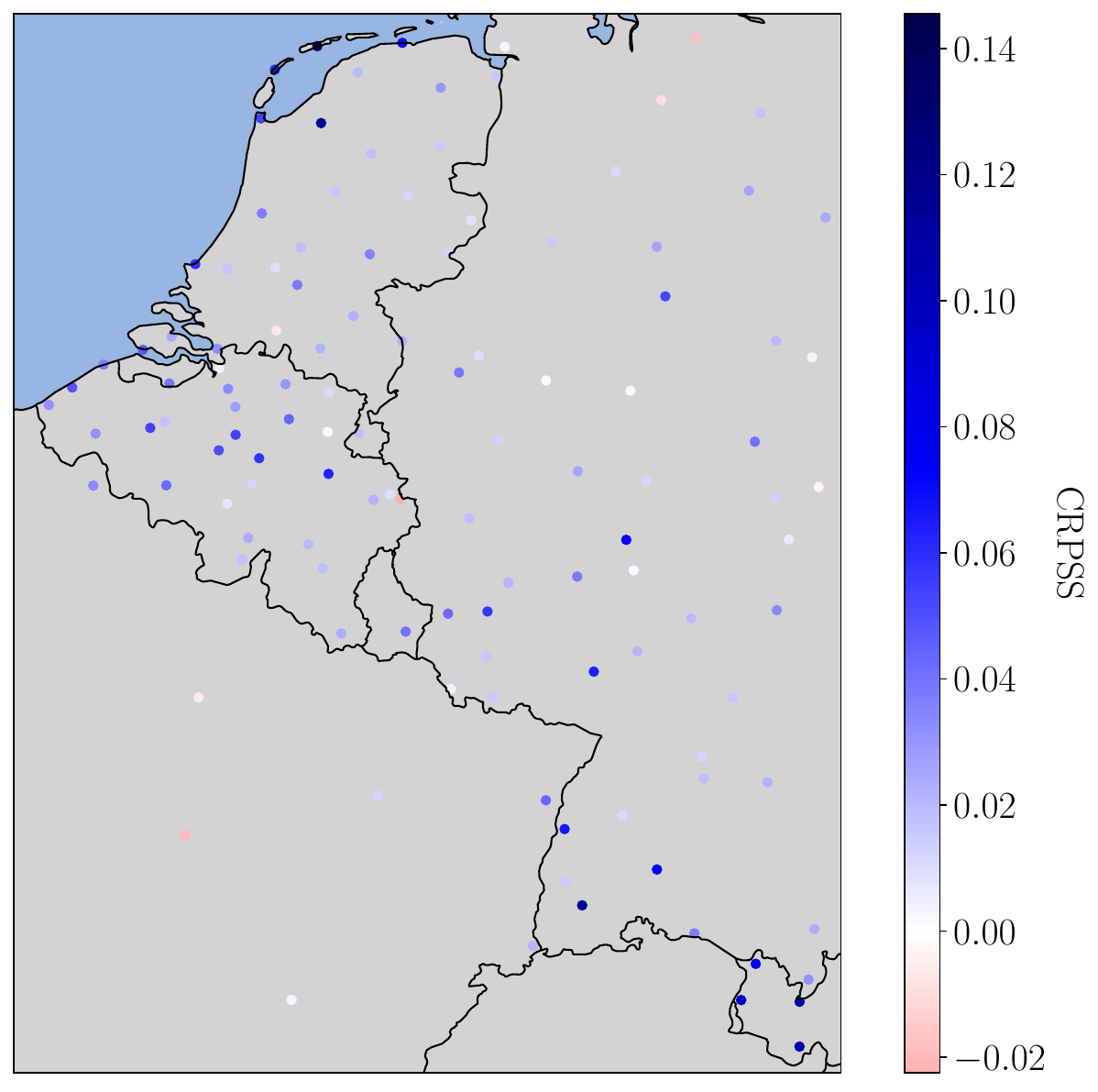}
		\vspace{-10pt}
		\caption{Station-specific improvement in terms of the CRPS of the GAT model over DRN, computed in terms of the CRPSS; where higher values indicate larger improvements by the GAT model.\vspace{-0.225in}}
		\phantomsection
		\label{fig:crpss}
	\end{center}
\end{figure}

In order to investigate local differences, \Cref{fig:crpss} shows the relative improvement in terms of the CRPS, i.e., the station-specific continuous ranked probability skill score, CRPSS ($=1-\text{CRPS}_{\text{GAT}}/\text{CRPS}_\text{DRN}$), where DRN serves as a reference method and $\text{CRPS}_{\text{GAT}}$ and $\text{CRPS}_\text{DRN}$ denote the corresponding mean CRPS at a  station.
The GAT model achieves improvements over DRN for almost all investigated stations, which range up to around 14\% in terms of the mean CRPS. 
While there is no clear geographical pattern, the improvements seem slightly larger at stations which are more centrally located within the graph.

Additional results on the 
calibration of the forecast distributions are available in the supplemental material.
To asses feature importance, we employ a permutation importance procedure with two stage feature shuffling, following \citet{hohlein_postprocessing_2024}.
The four most important features are all related to temperature variables from the NWP ensemble, followed by the station ID and the station altitude. Together these six features are responsible for roughly \SI{80}{\percent} of the total feature importance.
Details and graphical illustrations are provided in the supplemental material.

\section{Conclusion}
\label{sec:conclusions}

We propose a graph neural network architecture for ensemble post-processing which enables an improved information sharing across station locations and achieves consistent and significant improvements over a highly competitive NN-based post-processing model across lead times and forecasting tasks on a benchmark dataset.
Within the proposed GAT architecture, the attention mechanism is a specifically important component to achieving these improvements. 
Potential future extensions of the GAT model include extensions towards spatio-temporal GNNs \citep{li2021spatial} as well as other graph generation methods based on alternative, e.g.\ meteorologically motivated similarity-based distance metrics \citep{LerchBaran2017}.
Further, a more detailed investigation of station-specific benefits of the GAT model and their relation to meteorological factors such as weather patterns or seasonality provides an interesting avenue for further analysis. 

\clearpage 
\bibliography{references}
\bibliographystyle{icml2024}

\newpage
\appendix
\onecolumn
\section{Supplementary Material}
\setcounter{figure}{0}
\renewcommand{\thefigure}{A.\arabic{figure}} 
\renewcommand\theHfigure{Appendix.\thefigure}
\setcounter{table}{0}
\renewcommand{\thetable}{A.\arabic{table}} 
\renewcommand\theHtable{Appendix.\thetable}

The supplementary material is organized as follows. \Cref{sup:data} gives an overview of the data and features used, and \Cref{sup:res} provides additional details on the model architecture and training, as well as additional results.

\subsection{Data}
\label{sup:data}
\begin{figure}[ht]
	\vskip 0.2in
	\begin{center}
		\centerline{\includegraphics[width=0.5\columnwidth]{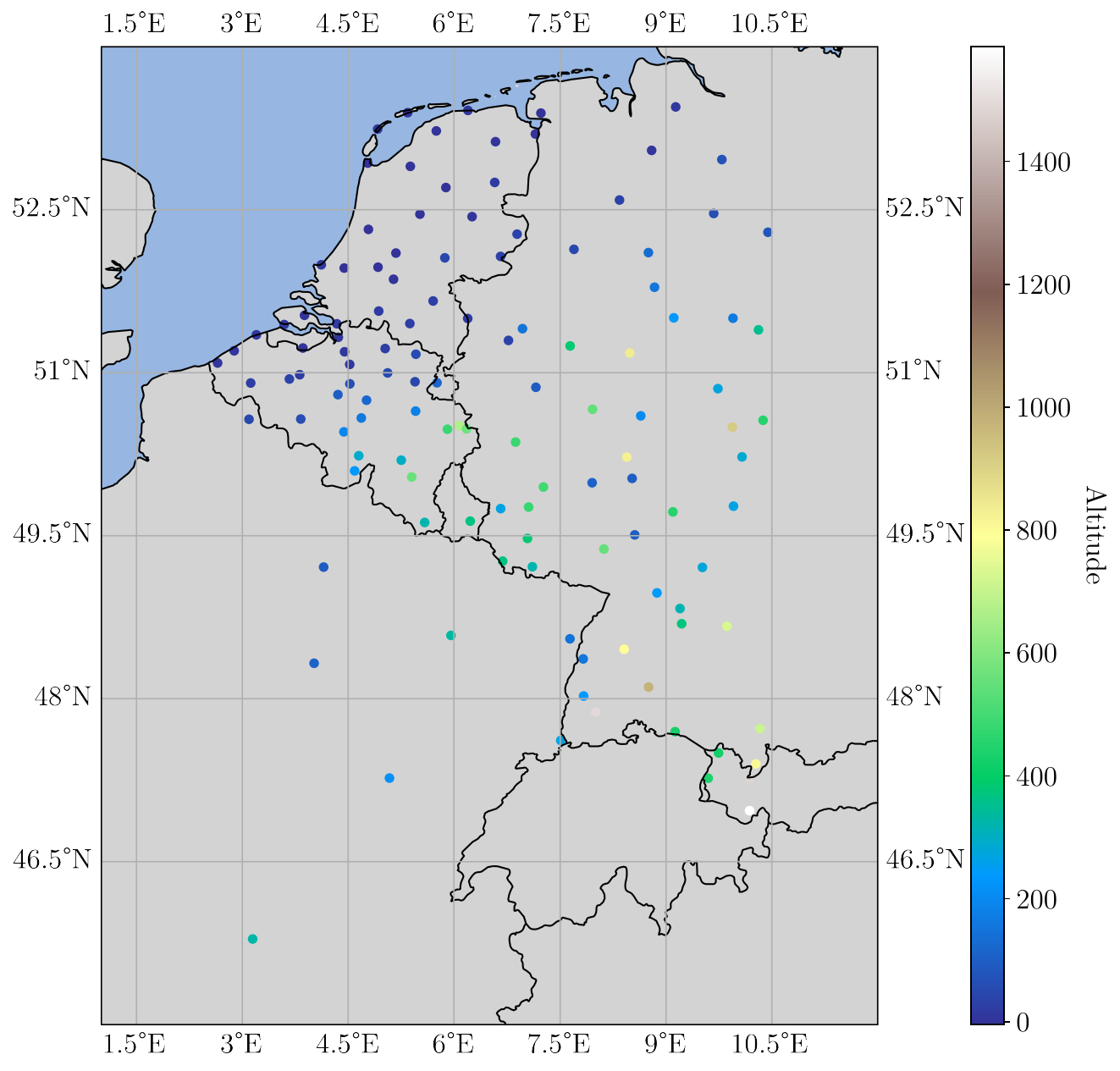}}
		\caption{Weather stations in the EUPPBench dataset with their corresponding altitude.}
		\phantomsection
		\label{fig:EUPP-Stations}
	\end{center}
	\vskip -0.2in
\end{figure}

The EUPPBench dataset \citep{demaeyer_euppbench_2023} includes (re)forecasts and observations of 2-m air temperature and additional auxiliary variables at lead times of 6 to \SI{120}{\hour} in \SI{6}{\hour} intervals for a total of 122 stations. The stations, along with their altitude, are shown in \Cref{fig:EUPP-Stations}.  The auxiliary variables are listed in \Cref{tab:axuiliary variables}, and station-specific information included in the dataset is listed in \Cref{tab:station specific}. The EUPPBench dataset is available through the CliMetLab API \cite{demaeyer_eupp-benchmarkclimetlab-eumetnet-postprocessing-benchmark_2024}

We focus on the lead times of \SIlist{24;72;120}{\hour}, and define a training, validation and testing datasets for the \textit{R2F} and \textit{R2R} tasks described in Section \ref{sec:data}.
An overview of the datasets and tasks is provided in Table \ref{tab:training data}.
Note that for the final model training, the valid set is used for training as well.

\begin{table}[htb]
	\caption{Sizes of the training, validation and test datasets in terms of the number of days for which a forecast is available. There is a forecast for 122 stations for each day, generated by either 11 or 51 ensemble members for the reforecasts and forecasts, respectively. RF\_Test and F\_Test denote the test datasets for the \textit{R2R} and \textit{R2F} task.}
	\label{tab:training data}
	\vskip 0.15in
	\begin{center}
		\begin{small}
			\begin{sc}
				\begin{tabular}{lccc}
					\toprule
					\textbf{Dataset} & \textbf{Size} & \textbf{Years} & \textbf{Reforecast?} \\
					\midrule
					Train & 2611 & 1997-2009 &\checkmark\\
					Valid & 836 & 2010-2013 &\checkmark\\
					RF\_Test & 733 & 2014-2017 &\checkmark\\
					F\_Test & 730 & 2017-2018 &$\times$ \\
					\bottomrule
				\end{tabular}
			\end{sc}
		\end{small}
	\end{center}
	\vskip -0.1in
\end{table}

\begin{table}[!htb]
	\centering
	\caption{Description of auxiliary variables, their corresponding units, full name, and levels which they were measured at \cite{demaeyer_euppbench_2023}. Temperature at \SI{2}{\meter} is the target variable of interest for our study.
		\textbf{Processed} indicates if the variable has been accumulated, averaged or filtered over the past \SI{6}{\hour}. Note that \textit{cin} is not used in the final dataset since the data is incomplete.}
	\phantomsection
	\label{tab:axuiliary variables}
	\vskip 0.15in
	\begin{center}
		\begin{small}
			\begin{threeparttable}
				\begin{sc}
					\begin{tabular}{lcp{7cm}cc}
						\toprule
						\textbf{Short name} & \textbf{Units} & \textbf{Full name} & \textbf{Levels} & \textbf{Processed?}\\
						\midrule
						t & \si{\kelvin} & Temperature & \SI{2}{\meter}, \SI{850}{\hecto\pascal} &\\
						mx2t6 & \si{\kelvin} & Max temperature & \SI{2}{\meter} & \checkmark \\
						mn2t6 & \si{\kelvin} & Min temperature & \SI{2}{\meter} & \checkmark \\
						z & \si{\meter\squared\per\second\squared} & Geopotential & \SI{500}{\hecto\pascal} & \\
						u & \si{\meter\per\second} & U component of wind & \SI{10}{\meter}, \SI{100}{\meter}, \SI{700}{\hecto\pascal} & \\
						v & \si{\meter\per\second} & V component of wind & \SI{10}{\meter}, \SI{100}{\meter}, \SI{700}{\hecto\pascal} & \\
						p10fg6 & \si{\meter\per\second} & Max wind gust & \SI{10}{\meter} & \checkmark \\
						q & \si{\kilogram\per\kilogram} & Specific humidity & \SI{700}{\hecto\pascal} & \\
						r & \% & Relative humidity & \SI{850}{\hecto\pascal} & \\
						cape & \si{\joule\per\kilogram} & Convective available potential energy & — & \\
						cin\tnote{1} & \si{\joule\per\kilogram} & Convective inhibition & — & \\
						tp6 & \si{\meter} & Total precipitation & — & \checkmark \\
						cp6 & \si{\meter} & Convective precipitation & — & \checkmark \\
						tcw & \si{\kilogram\per\square\meter} & Total column water & — & \\
						tcwv & \si{\kilogram\per\square\meter} & Total column water vapor & — & \\
						tcc & $\in[0,1]$ & Total cloud cover & — & \\
						vis & \si{\meter} & Visibility & — & \\
						sshf6 & \si{\joule\per\square\meter} & Surface sensible heat flux & — & \checkmark \\
						slhf6 & \si{\joule\per\square\meter} & Surface latent heat flux & — & \checkmark \\
						ssr6 & \si{\joule\per\square\meter} & Surface net shortwave (solar) radiation & — & \checkmark \\
						ssrd6 & \si{\joule\per\square\meter} & Surface net shortwave (solar) radiation downward & — & \checkmark \\
						str6 & \si{\joule\per\square\meter} & Surface net longwave (thermal) radiation & — & \checkmark \\
						strd6 & \si{\joule\per\square\meter} & Surface net longwave (thermal) radiation downward & — & \checkmark \\
						swv & \si{\cubic\meter\per\cubic\meter} & Volumetric soil water & L1: \SIrange{0}{7}{\centi \meter} & \\
						sd & \si{\meter} & Snow depth-water equivalent & — & \\
						st & \si{\kelvin} & Soil temperature & L1: \SIrange{0}{7}{\centi \meter} & \\
						\bottomrule
					\end{tabular}
				\end{sc}
				\begin{tablenotes}
					\item[1] Omitted in the final analysis due to missing data.
				\end{tablenotes}
			\end{threeparttable}
		\end{small}
	\end{center}
	\vskip -0.1in
\end{table}

\begin{table}[htb]
	\centering
	\caption{Further auxiliary variables, which are station specific except for yday, see \citet{schulz_machine_2022} for details.}
	\phantomsection
	\label{tab:station specific}
	\vskip 0.15in
	\begin{center}
		\begin{small}
			\begin{sc}
				\begin{tabular}{lcl}
					\toprule
					\textbf{Predictor} & \textbf{Type} & \textbf{Description} \\
					\midrule
					yday  & Temporal & Cosine and Sine transformed day of the year \\
					id    & — & Unique id assigned to each station\\
					lat   & Spatial & Latitude of the station \\
					lon   & Spatial & Longitude of the station \\
					alt   & Spatial & Altitude of the station \\
					orog  & Spatial & Difference of station altitude and model surface height of nearest grid point \\
					\bottomrule
				\end{tabular}
			\end{sc}
		\end{small}
	\end{center}
	\vskip -0.1in
\end{table}

\clearpage
\subsection{Additional results}
\label{sup:res}

\subsubsection{Details on hyperparameter optimization and model training}
\label{sec:hyper}

Following \citet{rasp_neural_2018}, a collection of 10 models is trained based on different random initalizations to address uncertainty during training and improve overall performance for all investigated post-processing models. 
The predictions, i.e., the distribution parameters obtained as the output of the resulting 10 models are averaged to generate the final prediction.
We use an early stopping algorithm to enable faster training; if the CRPS does not increase for 10 epochs, we revert to the best model iteration and stop training.
Model parameters are estimated using adaptive moment estimation with weight decay (\textit{AdamW}) \citep{loshchilov_decoupled_2019}.

\Cref{tab:hyperparameters} shows the results of a grid search for the GAT model. Note that $d_\text{max}$ was not included in the grid search, however preliminary testing showed that \SI{100}{\kilo\meter} delivered good results. 
Further, the DS and SMRY models were optimized using the same hyperparameter grid as for the GAT model. 
Similar to the approach for the graph based models, the DRN model was also optimized using a grid search of the relevant hyperparameters, see also the model descriptions in \citet{rasp_neural_2018, schulz_machine_2022}, and \citet{hohlein_postprocessing_2024}.
Training times ranged from a few minutes for the DRN to up to an hour for the GAT-based models on one NVIDIA P40 GPU. Note that in contrast to the computational costs of all post-processing methods are negligible compared to the costs of obtaining the raw forecasts by running ensembles of NWP models.

\begin{table}[htb]
	\centering
	\caption{Choice of hyperparameters of the GAT model. The column `Optimized?' indicates whether the hyperparameters were optimized based on the validation dataset using a grid search.}
	\phantomsection
	\label{tab:hyperparameters}
	\vskip 0.15in
	\begin{center}
		\begin{small}
			\begin{sc}
				\begin{tabular}{lcccc}
					\toprule
					\textbf{Parameter} & \textbf{\SI{24}{\hour}} & \textbf{\SI{72}{\hour}} & \textbf{\SI{120}{\hour}} & \textbf{Optimized?}\\
					\midrule
					maximal distance ($d_{\text{max}}$) & \SI{100}{\kilo\meter} & \SI{100}{\kilo\meter} & \SI{100}{\kilo\meter}\\[0.5em]
					batch size  & 8 & 8 & 8 \\
					training epochs  & 31 & 42 & 35 & \checkmark\\
					learning rate   & 0.0002 & 0.0001 & 0.0005 & \checkmark\\[0.5em]
					embedding dimension & 20 & 20 & 20\\
					hidden channels (GNN) & 265 & 128 & 64 & \checkmark\\
					GNN Layers   & 2 & 2 & 1 & \checkmark\\
					attention heads & 8 & 8 & 8 & \checkmark\\[0.5em]
					deep set layers (in) & 3 & 3 & 3\\
					deep set layers (out) & 2 & 2 & 2\\
					deep set hidden channels & \multicolumn{3}{c}{same as ``hidden channels (GNN)''}& (\checkmark)\\[0.5em]
					\bottomrule
				\end{tabular}
			\end{sc}
		\end{small}
	\end{center}
	\vskip -0.1in
\end{table}

\subsubsection{Additional results}
\label{sup:sig}

To compare the different post-processing models, we report the average CRPS for the two tasks (\textit{R2R} and \textit{R2F}) and all models in \Cref{tab:scores_full}, along with the average length of the prediction interval (PI length) based on a nominal level of $N-1/N+1$, where $N$ is the number of ensemble members. This evaluates to \SIlist{96.15;83.33}{\percent} for the \textit{R2F} and \textit{R2R} task, respectively. 
Overall, qualitatively similar results are obtained for the two tasks, with similar rankings and relative improvements of the GAT model over the alternative specification of GNN models and the DRN model.

\begin{table*}[htb]
	\caption{Scores for the reforecast to reforecast (\textit{R2R}) and reforecast to forecast (\textit{R2F}) tasks. Scores are calculated per lead time, with the best CRPS scores highlighted in \textbf{bold}.}
	\label{tab:scores_full}
	\vskip 0.15in
	\begin{center}
		\begin{small}
			\begin{sc}
				\begin{tabular}{lccccccccc}
					\toprule
					\textbf{Lead time} & \multicolumn{3}{c}{\SI{24}{\hour}} & \multicolumn{3}{c}{\SI{72}{\hour}} & \multicolumn{3}{c}{\SI{120}{\hour}}\\
					
					\cmidrule(lr){1-1}
					\cmidrule(lr){2-4}
					\cmidrule(lr){5-7}
					\cmidrule(lr){8-10}
					
					\multicolumn{1}{c}{\textbf{Method}} & CRPS & PI length & PI cover & CRPS & PI length & PI cover & CRPS & PI length & PI cover\\
					
					\midrule
					\multicolumn{10}{l}{\textit{R2R}} \\[0.5em]
					\quad ENS & 1.20 & 1.82 & 38.94 & 1.28 & 3.29 & 54.98 & 1.54 & 4.88 & 61.60\\
					\quad DRN & 0.65 & 2.79 & 78.38 & 0.86 & 3.89 & 80.32 & 1.19 & 5.27 & 79.53\\[0.5em]
					\quad SMRY & 0.66 & 2.91 & 79.73 & 0.87 & 4.04 & 82.37 & 1.18 & 5.48 & 81.51\\
					\quad DS & 0.64 & 2.92 & 81.11 & 0.87 & 2.90 & 68.53 & 1.25  & 3.01 & 54.97\\
					\quad GAT & \textbf{0.63} & 2.75 & 79.39 & \textbf{0.85} & 3.90 & 81.66 & \textbf{1.17} & 5.47 & 82.32\\
					\midrule
					\multicolumn{10}{l}{\textit{R2F}} \\[0.5em]
					\quad ENS & 1.12 & 2.66 & 56.06 & 1.18 & 4.72 & 72.90 & 1.38 & 7.14 & 81.16 \\
					\quad DRN & 0.61 & 4.26 & 94.87 & 0.79 & 5.90 & 96.37 & 1.11 & 7.99 & 95.82\\[0.5em]
					\quad SMRY & 0.62 & 4.45 & 95.53 & 0.79 & 6.17 & 97.01 & 1.10 & 8.31 & 96.64 \\
					\quad DS & 0.61 & 4.41 & 95.72 & \textbf{0.78} & 4.43 & 89.87 & 1.14 & 4.56 & 77.79\\
					\quad GAT & \textbf{0.60} & 4.16 & 95.04 & \textbf{0.78} & 5.93 & 96.42 & \textbf{1.09} & 8.27 & 96.80\\
					\bottomrule
				\end{tabular}
			\end{sc}
		\end{small}
	\end{center}
	\vskip -0.2in
\end{table*}

To assess the statistical significance of score differences, we use  Diebold-Mariano tests \cite{DieboldMariano1995} of equal predictive performance.
The test is conducted for each combination of two models and  separately for the considered lead times, with the null hypothesis of equal predictive performance at a given station. The test statistic is  
\begin{equation*}
t = \sqrt{n}\frac{\bar{S}^F_n-\bar{S}^G_n}{\hat{\sigma}_n},\; \text{where}\; \hat{\sigma}_n=\frac{1}{n}\sum_{i=1}^{n}\left(S(F_i,y_i)-S(G_i,y_i)\right)^2,
\end{equation*}
and $\bar{S}^F$ and $\bar{S}^G$ denote the corresponding mean scores for a fixed observation station and lead time for the two models' forecast distributions $F$ and $G$ and a corresponding test dataset of size $n$.
Under the assumption of equal predictive performance, the distribution of $t$ approximately follows a standard Gaussian distribution.
In order to account for multiple testing, the Benjamini-Hochberg correction is applied \cite{benjamini_controlling_1995}, which corresponds to sorting the p-values of the per-station tests in ascending order and selecting the corrected significance level as
\begin{equation*}
p^*= \max(p_i \,\vert\, p_i\leq\frac{\alpha i}{M}).
\end{equation*}
For all p-values smaller or equal to $p^*$ the null hypothesis is rejected. 
Results are reported in \Cref{tab:scores_significance} and indicate that the GAT models' scores tend to be significantly better those of DRN at up to 38\% of the stations, while the null hypothesis is never rejected in favor of the DRN model. 
For longer lead times, the fraction of stations with significant score differences tends to decrease, and overall, qualitatively similar results can be observed for the two tasks.

\begin{table*}[htb]
	\caption{Percentage of combinations of stations showing statistically significant differences in terms of the CRPS after applying the Benjamini–Hochberg correction at a nominal level of 0.05. Two-sided test were conducted, the table shows the ratio of stations for which the null hypothesis of equal predictive performance was rejected in favor of the model in the row, while comparing it to the model in the column.}
	\label{tab:scores_significance}
	\vskip 0.15in
	\begin{center}
		\begin{small}
			\newcolumntype{R}{>{\raggedleft\arraybackslash}X}
			\begin{tabularx}{\textwidth}{ lRRRR c RRRR c RRRR }
				\toprule
				\textbf{Lead time} & \multicolumn{4}{c}{\SI{24}{\hour}} & \multicolumn{1}{c}{} & \multicolumn{4}{c}{\SI{72}{\hour}} & \multicolumn{1}{c}{} & \multicolumn{4}{c}{\SI{120}{\hour}}\\
				\cmidrule(lr){1-1}
				\cmidrule(lr){2-5}
				\cmidrule(lr){7-10}
				\cmidrule(lr){12-15}
				
				\textbf{Method} & DRN & SMRY & DS & GAT && DRN & SMRY & DS & GAT && DRN & SMRY & DS & GAT\\
				\midrule
				\multicolumn{10}{l}{\textit{R2R}} \\[0.5em]
				\quad DRN & — & 29.2 & 0.0 & 0.0 && — & 11.7 & 0.8 & 0.0 && — & 1.7 & 84.2 & 0.0\\
				\quad SMRY & 2.5 & — & 0.0 & 0.0 && 5.8 & — & 0.0 & 0.0 && 5.0 & — & 89.2 & 0.0\\
				\quad DS & 18.3 & 36.7 & — & 1.7 && 3.3 & 2.5 & — & 0.0 && 0.8 & 0.8 & — & 0.0\\
				\quad GAT & 30.0 & 70.0 & 23.3 & — && 20.8 & 45.8 & 51.7 & — && 16.7 & 22.5 & 97.5 & —\\
				\midrule
				\multicolumn{10}{l}{\textit{R2F}} \\[0.5em]
				\quad DRN & — & 30.0 & 0.0 & 0.0 && — & 22.5 & 0.0 & 1.7 && — & 8.3 & 50.8 & 0.0\\
				\quad SMRY & 7.5 & — & 5.0 & 0.0 && 10.0 & — & 0.0 & 3.3 && 10.0 & — & 55.0 & 0.0 \\
				\quad DS & 10.8 & 40.8 & — & 2.5 && 20.0 & 28.3 & — & 10.0 && 1.7 & 3.3 & — & 0.0\\
				\quad GAT & 38.3 & 50.0 & 24.2 & — && 14.2 & 35.8 & 0.0 & — && 10.8 & 11.7 & 72.5 & —\\
				\bottomrule
			\end{tabularx}
		\end{small}
	\end{center}
	\vskip -0.1in
\end{table*}


\subsubsection{PIT Histograms}
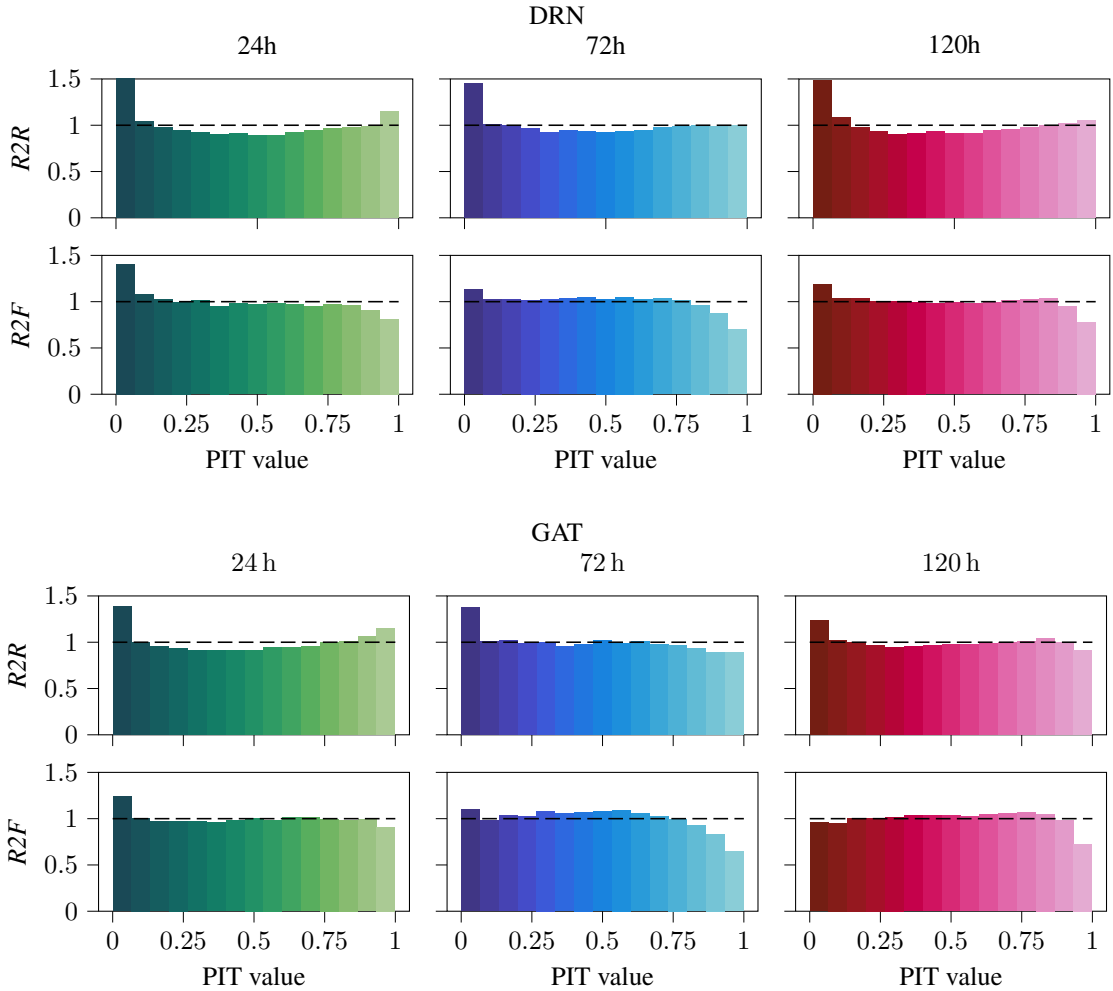
\begin{figure}
	\vskip 0.2in
	\begin{center}
		DRN\\
		\begin{tikzpicture}
		
		\definecolor{brown1492431}{RGB}{149,24,31}
		\definecolor{crimson197175}{RGB}{197,1,75}
		\definecolor{crimson2081996}{RGB}{208,19,96}
		\definecolor{darkgray176}{RGB}{176,176,176}
		\definecolor{darkseagreen11418199}{RGB}{114,181,99}
		\definecolor{darkseagreen136187112}{RGB}{136,187,112}
		\definecolor{darkseagreen154194130}{RGB}{154,194,130}
		\definecolor{darkseagreen170202148}{RGB}{170,202,148}
		\definecolor{darkslateblue6454133}{RGB}{64,54,133}
		\definecolor{darkslateblue6860156}{RGB}{68,60,156}
		\definecolor{darkslateblue7067179}{RGB}{70,67,179}
		\definecolor{darkslategray1910399}{RGB}{19,103,99}
		\definecolor{darkslategray219395}{RGB}{21,93,95}
		\definecolor{darkslategray248391}{RGB}{24,83,91}
		\definecolor{darkslategray267386}{RGB}{26,73,86}
		\definecolor{dodgerblue25131222}{RGB}{25,131,222}
		\definecolor{dodgerblue29143220}{RGB}{29,143,220}
		\definecolor{dodgerblue41155217}{RGB}{41,155,217}
		\definecolor{firebrick1661641}{RGB}{166,16,41}
		\definecolor{firebrick181555}{RGB}{181,5,55}
		\definecolor{maroon1163019}{RGB}{116,30,19}
		\definecolor{maroon1322824}{RGB}{132,28,24}
		\definecolor{mediumseagreen6416497}{RGB}{64,164,97}
		\definecolor{mediumseagreen8817494}{RGB}{88,174,94}
		\definecolor{mediumturquoise59167215}{RGB}{59,167,215}
		\definecolor{mediumturquoise77177214}{RGB}{77,177,214}
		\definecolor{mediumturquoise97187213}{RGB}{97,187,213}
		\definecolor{mediumvioletred21541117}{RGB}{215,41,117}
		\definecolor{mediumvioletred22062137}{RGB}{220,62,137}
		\definecolor{orchid225122182}{RGB}{225,122,182}
		\definecolor{orchid226139192}{RGB}{226,139,192}
		\definecolor{palevioletred22382154}{RGB}{223,82,154}
		\definecolor{palevioletred225104170}{RGB}{225,104,170}
		\definecolor{plum227155202}{RGB}{227,155,202}
		\definecolor{plum228171210}{RGB}{228,171,210}
		\definecolor{royalblue33118223}{RGB}{33,118,223}
		\definecolor{royalblue46104223}{RGB}{46,104,223}
		\definecolor{royalblue6089216}{RGB}{60,89,216}
		\definecolor{royalblue6876201}{RGB}{68,76,201}
		\definecolor{seagreen24135103}{RGB}{24,135,103}
		\definecolor{seagreen34145102}{RGB}{34,145,102}
		\definecolor{seagreen47155100}{RGB}{47,155,100}
		\definecolor{skyblue117196214}{RGB}{117,196,214}
		\definecolor{skyblue138205215}{RGB}{138,205,215}
		\definecolor{teal18114101}{RGB}{18,114,101}
		\definecolor{teal19125102}{RGB}{19,125,102}
		
		\begin{groupplot}[group style={group size=3 by 2,horizontal sep=0.5cm,vertical sep=0.5cm },height=\textwidth/5,width=\textwidth/3]
		\nextgroupplot[
		scaled x ticks=manual:{}{\pgfmathparse{#1}},
		tick align=outside,
		tick pos=left,
		title={24h},
		x grid style={darkgray176},
		xmin=-0.05, xmax=1.05,
		xtick style={color=black},
		xticklabels={},
		y grid style={darkgray176},
		ylabel={\textit{R2R}},
		ymin=0, ymax=1.5,
		ytick style={color=black},
		ytick={0,0.5,1,1.5},
		]
		\draw[draw=none,fill=darkslategray267386] (axis cs:-6.93889390390723e-18,0) rectangle (axis cs:0.0666666666666667,1.5659132194348);
		\draw[draw=none,fill=darkslategray248391] (axis cs:0.0666666666666667,0) rectangle (axis cs:0.133333333333333,1.04129779944354);
		\draw[draw=none,fill=darkslategray219395] (axis cs:0.133333333333333,0) rectangle (axis cs:0.2,0.975591528892363);
		\draw[draw=none,fill=darkslategray1910399] (axis cs:0.2,0) rectangle (axis cs:0.266666666666667,0.942307250109223);
		\draw[draw=none,fill=teal18114101] (axis cs:0.266666666666667,0) rectangle (axis cs:0.333333333333333,0.923509393179885);
		\draw[draw=none,fill=teal19125102] (axis cs:0.333333333333333,0) rectangle (axis cs:0.4,0.89815815493573);
		\draw[draw=none,fill=seagreen24135103] (axis cs:0.4,0) rectangle (axis cs:0.466666666666667,0.907643312101911);
		\draw[draw=none,fill=seagreen34145102] (axis cs:0.466666666666667,0) rectangle (axis cs:0.533333333333333,0.893674262457173);
		\draw[draw=none,fill=seagreen47155100] (axis cs:0.533333333333333,0) rectangle (axis cs:0.6,0.892811975442066);
		\draw[draw=none,fill=mediumseagreen6416497] (axis cs:0.6,0) rectangle (axis cs:0.666666666666667,0.9276483708524);
		\draw[draw=none,fill=mediumseagreen8817494] (axis cs:0.666666666666667,0) rectangle (axis cs:0.733333333333333,0.945411483363609);
		\draw[draw=none,fill=darkseagreen11418199] (axis cs:0.733333333333333,0) rectangle (axis cs:0.8,0.963347053277839);
		\draw[draw=none,fill=darkseagreen136187112] (axis cs:0.8,0) rectangle (axis cs:0.866666666666667,0.978350847340707);
		\draw[draw=none,fill=darkseagreen154194130] (axis cs:0.866666666666667,0) rectangle (axis cs:0.933333333333333,0.991285152567316);
		\draw[draw=none,fill=darkseagreen170202148] (axis cs:0.933333333333333,0) rectangle (axis cs:1,1.15305019660144);
		\path [draw=black, semithick, dash pattern=on 5.55pt off 2.4pt]
		(axis cs:0,1)
		--(axis cs:1,1);

		\nextgroupplot[
		scaled x ticks=manual:{}{\pgfmathparse{#1}},
		scaled y ticks=manual:{}{\pgfmathparse{#1}},
		tick align=outside,
		tick pos=left,
		title={72h},
		x grid style={darkgray176},
		xmin=-0.05, xmax=1.05,
		xtick style={color=black},
		xticklabels={},
		y grid style={darkgray176},
		ymin=0, ymax=1.5,
		ytick style={color=black},
		yticklabels={}
		]
		\draw[draw=none,fill=darkslateblue6454133] (axis cs:1.80411241501588e-16,0) rectangle (axis cs:0.0666666666616433,1.44907040126817);
		\draw[draw=none,fill=darkslateblue6860156] (axis cs:0.0666666666616433,0) rectangle (axis cs:0.133333333323286,1.01479769591311);
		\draw[draw=none,fill=darkslateblue7067179] (axis cs:0.133333333323287,0) rectangle (axis cs:0.19999999998493,1.00031044124677);
		\draw[draw=none,fill=royalblue6876201] (axis cs:0.19999999998493,0) rectangle (axis cs:0.266666666646573,0.9642647718984);
		\draw[draw=none,fill=royalblue6089216] (axis cs:0.266666666646573,0) rectangle (axis cs:0.333333333308216,0.924942223518354);
		\draw[draw=none,fill=royalblue46104223] (axis cs:0.333333333308216,0) rectangle (axis cs:0.399999999969859,0.944603497708376);
		\draw[draw=none,fill=royalblue33118223] (axis cs:0.399999999969859,0) rectangle (axis cs:0.466666666631502,0.93546272988319);
		\draw[draw=none,fill=dodgerblue25131222] (axis cs:0.466666666631502,0) rectangle (axis cs:0.533333333293145,0.924079886931071);
		\draw[draw=none,fill=dodgerblue29143220] (axis cs:0.533333333293145,0) rectangle (axis cs:0.599999999954788,0.933220654756258);
		\draw[draw=none,fill=dodgerblue41155217] (axis cs:0.599999999954788,0) rectangle (axis cs:0.666666666616432,0.950294919184436);
		\draw[draw=none,fill=mediumturquoise59167215] (axis cs:0.666666666616432,0) rectangle (axis cs:0.733333333278075,0.972888137771215);
		\draw[draw=none,fill=mediumturquoise77177214] (axis cs:0.733333333278075,0) rectangle (axis cs:0.799999999939718,0.989272532929569);
		\draw[draw=none,fill=mediumturquoise97187213] (axis cs:0.799999999939718,0) rectangle (axis cs:0.866666666601361,1.00341485296099);
		\draw[draw=none,fill=skyblue117196214] (axis cs:0.866666666601361,0) rectangle (axis cs:0.933333333263004,0.988927598294654);
		\draw[draw=none,fill=skyblue138205215] (axis cs:0.933333333263004,0) rectangle (axis cs:0.999999999924647,1.00444965686573);
		\path [draw=black, semithick, dash pattern=on 5.55pt off 2.4pt]
		(axis cs:0,1)
		--(axis cs:1,1);

		\nextgroupplot[
		scaled x ticks=manual:{}{\pgfmathparse{#1}},
		scaled y ticks=manual:{}{\pgfmathparse{#1}},
		tick align=outside,
		tick pos=left,
		title={120h},
		x grid style={darkgray176},
		xmin=-0.05, xmax=1.05,
		xtick style={color=black},
		xticklabels={},
		y grid style={darkgray176},
		ymin=0, ymax=1.5,
		ytick style={color=black},
		yticklabels={}
		]
		\draw[draw=none,fill=maroon1163019] (axis cs:1.31326893804129e-12,0) rectangle (axis cs:0.0666666652243289,1.48899598477812);
		\draw[draw=none,fill=maroon1322824] (axis cs:0.066666665224329,0) rectangle (axis cs:0.133333330447345,1.0829845694917);
		\draw[draw=none,fill=brown1492431] (axis cs:0.133333330447345,0) rectangle (axis cs:0.19999999567036,0.975531251002556);
		\draw[draw=none,fill=firebrick1661641] (axis cs:0.19999999567036,0) rectangle (axis cs:0.266666660893376,0.934654145895129);
		\draw[draw=none,fill=firebrick181555] (axis cs:0.266666660893376,0) rectangle (axis cs:0.333333326116392,0.898778880653168);
		\draw[draw=none,fill=crimson197175] (axis cs:0.333333326116392,0) rectangle (axis cs:0.399999991339407,0.911369718935203);
		\draw[draw=none,fill=crimson2081996] (axis cs:0.399999991339407,0) rectangle (axis cs:0.466666656562423,0.931377078397066);
		\draw[draw=none,fill=mediumvioletred21541117] (axis cs:0.466666656562423,0) rectangle (axis cs:0.533333321785439,0.916544036037409);
		\draw[draw=none,fill=mediumvioletred22062137] (axis cs:0.533333321785439,0) rectangle (axis cs:0.599999987008454,0.913094491302605);
		\draw[draw=none,fill=palevioletred22382154] (axis cs:0.599999987008454,0) rectangle (axis cs:0.66666665223147,0.941380758127997);
		\draw[draw=none,fill=palevioletred225104170] (axis cs:0.66666665223147,0) rectangle (axis cs:0.733333317454486,0.951039483385448);
		\draw[draw=none,fill=orchid225122182] (axis cs:0.733333317454486,0) rectangle (axis cs:0.799999982677501,0.975531251002556);
		\draw[draw=none,fill=orchid226139192] (axis cs:0.799999982677501,0) rectangle (axis cs:0.866666647900517,1.00123035927685);
		\draw[draw=none,fill=plum227155202] (axis cs:0.866666647900517,0) rectangle (axis cs:0.933333313123533,1.02606708136743);
		\draw[draw=none,fill=plum228171210] (axis cs:0.933333313123533,0) rectangle (axis cs:0.999999978346548,1.05142123516824);
		\path [draw=black, semithick, dash pattern=on 5.55pt off 2.4pt]
		(axis cs:0,1)
		--(axis cs:1,1);

		\nextgroupplot[
		tick align=outside,
		tick pos=left,
		x grid style={darkgray176},
		xlabel={PIT value},
		xmin=-0.05, xmax=1.05,
		xtick style={color=black},
		xtick={0,0.25,0.5,0.75,1},
		y grid style={darkgray176},
		ylabel={\textit{R2F}},
		ymin=0, ymax=1.5,
		ytick style={color=black},
		ytick={0,0.5,1,1.5},
		]
		\draw[draw=none,fill=darkslategray267386] (axis cs:6.93889390390723e-18,0) rectangle (axis cs:0.0666657776175501,1.40729551906913);
		\draw[draw=none,fill=darkslategray248391] (axis cs:0.0666657776175501,0) rectangle (axis cs:0.1333315552351,1.08201622792032);
		\draw[draw=none,fill=darkslategray219395] (axis cs:0.1333315552351,0) rectangle (axis cs:0.19999733285265,1.02702483260648);
		\draw[draw=none,fill=darkslategray1910399] (axis cs:0.19999733285265,0) rectangle (axis cs:0.266663110470201,0.995897627711858);
		\draw[draw=none,fill=teal18114101] (axis cs:0.266663110470201,0) rectangle (axis cs:0.333328888087751,1.01076951449484);
		\draw[draw=none,fill=teal19125102] (axis cs:0.333328888087751,0) rectangle (axis cs:0.399994665705301,0.95560519026493);
		\draw[draw=none,fill=seagreen24135103] (axis cs:0.399994665705301,0) rectangle (axis cs:0.466660443322851,0.981890385509278);
		\draw[draw=none,fill=seagreen34145102] (axis cs:0.466660443322851,0) rectangle (axis cs:0.533326220940401,0.971687579460485);
		\draw[draw=none,fill=seagreen47155100] (axis cs:0.533326220940401,0) rectangle (axis cs:0.599991998557951,0.987769968656041);
		\draw[draw=none,fill=mediumseagreen6416497] (axis cs:0.599991998557951,0) rectangle (axis cs:0.666657776175501,0.974973228866028);
		\draw[draw=none,fill=mediumseagreen8817494] (axis cs:0.666657776175501,0) rectangle (axis cs:0.733323553793052,0.954740545684524);
		\draw[draw=none,fill=darkseagreen11418199] (axis cs:0.733323553793052,0) rectangle (axis cs:0.799989331410602,0.967191427642373);
		\draw[draw=none,fill=darkseagreen136187112] (axis cs:0.799989331410602,0) rectangle (axis cs:0.866655109028152,0.962522346908178);
		\draw[draw=none,fill=darkseagreen154194130] (axis cs:0.866655109028152,0) rectangle (axis cs:0.933320886645702,0.907703880510427);
		\draw[draw=none,fill=darkseagreen170202148] (axis cs:0.933320886645702,0) rectangle (axis cs:0.999986664263252,0.813111763413987);
		\path [draw=black, semithick, dash pattern=on 5.55pt off 2.4pt]
		(axis cs:0,1)
		--(axis cs:1,1);

		\nextgroupplot[
		scaled y ticks=manual:{}{\pgfmathparse{#1}},
		tick align=outside,
		tick pos=left,
		x grid style={darkgray176},
		xlabel={PIT value},
		xmin=-0.05, xmax=1.05,
		xtick style={color=black},
		xtick={0,0.25,0.5,0.75,1},
		y grid style={darkgray176},
		ymin=0, ymax=1.5,
		ytick style={color=black},
		yticklabels={},
		]
		\draw[draw=none,fill=darkslateblue6454133] (axis cs:0,0) rectangle (axis cs:0.0666660132380165,1.13302625364826);
		\draw[draw=none,fill=darkslateblue6860156] (axis cs:0.0666660132380165,0) rectangle (axis cs:0.133332026476033,1.02684827444496);
		\draw[draw=none,fill=darkslateblue7067179] (axis cs:0.133332026476033,0) rectangle (axis cs:0.19999803971405,1.02477313478627);
		\draw[draw=none,fill=royalblue6876201] (axis cs:0.19999803971405,0) rectangle (axis cs:0.266664052952066,1.01975821394442);
		\draw[draw=none,fill=royalblue6089216] (axis cs:0.266664052952066,0) rectangle (axis cs:0.333330066190083,1.02148749699333);
		\draw[draw=none,fill=royalblue46104223] (axis cs:0.333330066190083,0) rectangle (axis cs:0.399996079428099,1.0360134746042);
		\draw[draw=none,fill=royalblue33118223] (axis cs:0.399996079428099,0) rectangle (axis cs:0.466662092666116,1.04431403323899);
		\draw[draw=none,fill=dodgerblue25131222] (axis cs:0.466662092666116,0) rectangle (axis cs:0.533328105904132,1.03134441037214);
		\draw[draw=none,fill=dodgerblue29143220] (axis cs:0.533328105904132,0) rectangle (axis cs:0.599994119142149,1.04915602577595);
		\draw[draw=none,fill=dodgerblue41155217] (axis cs:0.599994119142149,0) rectangle (axis cs:0.666660132380165,1.03047976884768);
		\draw[draw=none,fill=mediumturquoise59167215] (axis cs:0.666660132380165,0) rectangle (axis cs:0.733326145618182,1.03514883307975);
		\draw[draw=none,fill=mediumturquoise77177214] (axis cs:0.733326145618182,0) rectangle (axis cs:0.799992158856198,1.01214936852919);
		\draw[draw=none,fill=mediumturquoise97187213] (axis cs:0.799992158856198,0) rectangle (axis cs:0.866658172094215,0.963902371464502);
		\draw[draw=none,fill=skyblue117196214] (axis cs:0.866658172094215,0) rectangle (axis cs:0.933324185332231,0.870521086823163);
		\draw[draw=none,fill=skyblue138205215] (axis cs:0.933324185332231,0) rectangle (axis cs:0.999990198570248,0.701224276334511);
		\path [draw=black, semithick, dash pattern=on 5.55pt off 2.4pt]
		(axis cs:0,1)
		--(axis cs:1,1);

		\nextgroupplot[
		scaled y ticks=manual:{}{\pgfmathparse{#1}},
		tick align=outside,
		tick pos=left,
		x grid style={darkgray176},
		xlabel={PIT value},
		xmin=-0.05, xmax=1.05,
		xtick style={color=black},
		xtick={0,0.25,0.5,0.75,1},
		y grid style={darkgray176},
		ymin=0, ymax=1.5,
		ytick style={color=black},
		yticklabels={},
		]
		\draw[draw=none,fill=maroon1163019] (axis cs:1.46792300537157e-13,0) rectangle (axis cs:0.0666665752862938,1.18973670721494);
		\draw[draw=none,fill=maroon1322824] (axis cs:0.0666665752862938,0) rectangle (axis cs:0.133333150572441,1.03375669123996);
		\draw[draw=none,fill=brown1492431] (axis cs:0.133333150572441,0) rectangle (axis cs:0.199999725858588,1.03617766709766);
		\draw[draw=none,fill=firebrick1661641] (axis cs:0.199999725858588,0) rectangle (axis cs:0.266666301144735,1.00418620040656);
		\draw[draw=none,fill=firebrick181555] (axis cs:0.266666301144735,0) rectangle (axis cs:0.333332876430882,1.00556961518239);
		\draw[draw=none,fill=crimson197175] (axis cs:0.333332876430882,0) rectangle (axis cs:0.399999451717029,0.99519400436366);
		\draw[draw=none,fill=crimson2081996] (axis cs:0.399999451717029,0) rectangle (axis cs:0.466666027003176,0.98239741768722);
		\draw[draw=none,fill=mediumvioletred21541117] (axis cs:0.466666027003176,0) rectangle (axis cs:0.533332602289323,0.98862278417846);
		\draw[draw=none,fill=mediumvioletred22062137] (axis cs:0.533332602289323,0) rectangle (axis cs:0.59999917757547,0.979457661288578);
		\draw[draw=none,fill=palevioletred22382154] (axis cs:0.59999917757547,0) rectangle (axis cs:0.666665752861617,0.993291809046891);
		\draw[draw=none,fill=palevioletred225104170] (axis cs:0.666665752861617,0) rectangle (axis cs:0.733332328147764,1.0112762011327);
		\draw[draw=none,fill=orchid225122182] (axis cs:0.733332328147764,0) rectangle (axis cs:0.799998903433911,1.02182473879841);
		\draw[draw=none,fill=orchid226139192] (axis cs:0.799998903433911,0) rectangle (axis cs:0.866665478720058,1.03220034961715);
		\draw[draw=none,fill=plum227155202] (axis cs:0.866665478720058,0) rectangle (axis cs:0.933332054006205,0.945736926127691);
		\draw[draw=none,fill=plum228171210] (axis cs:0.933332054006205,0) rectangle (axis cs:0.999998629292352,0.780591787262824);
		\path [draw=black, semithick, dash pattern=on 5.55pt off 2.4pt]
		(axis cs:0,1)
		--(axis cs:1,1);
		
		\end{groupplot}
		
		\end{tikzpicture}
		\bigbreak
		GAT\\
		\begin{tikzpicture}
		
		\definecolor{brown1492431}{RGB}{149,24,31}
		\definecolor{crimson197175}{RGB}{197,1,75}
		\definecolor{crimson2081996}{RGB}{208,19,96}
		\definecolor{darkgray176}{RGB}{176,176,176}
		\definecolor{darkseagreen11418199}{RGB}{114,181,99}
		\definecolor{darkseagreen136187112}{RGB}{136,187,112}
		\definecolor{darkseagreen154194130}{RGB}{154,194,130}
		\definecolor{darkseagreen170202148}{RGB}{170,202,148}
		\definecolor{darkslateblue6454133}{RGB}{64,54,133}
		\definecolor{darkslateblue6860156}{RGB}{68,60,156}
		\definecolor{darkslateblue7067179}{RGB}{70,67,179}
		\definecolor{darkslategray1910399}{RGB}{19,103,99}
		\definecolor{darkslategray219395}{RGB}{21,93,95}
		\definecolor{darkslategray248391}{RGB}{24,83,91}
		\definecolor{darkslategray267386}{RGB}{26,73,86}
		\definecolor{dodgerblue25131222}{RGB}{25,131,222}
		\definecolor{dodgerblue29143220}{RGB}{29,143,220}
		\definecolor{dodgerblue41155217}{RGB}{41,155,217}
		\definecolor{firebrick1661641}{RGB}{166,16,41}
		\definecolor{firebrick181555}{RGB}{181,5,55}
		\definecolor{maroon1163019}{RGB}{116,30,19}
		\definecolor{maroon1322824}{RGB}{132,28,24}
		\definecolor{mediumseagreen6416497}{RGB}{64,164,97}
		\definecolor{mediumseagreen8817494}{RGB}{88,174,94}
		\definecolor{mediumturquoise59167215}{RGB}{59,167,215}
		\definecolor{mediumturquoise77177214}{RGB}{77,177,214}
		\definecolor{mediumturquoise97187213}{RGB}{97,187,213}
		\definecolor{mediumvioletred21541117}{RGB}{215,41,117}
		\definecolor{mediumvioletred22062137}{RGB}{220,62,137}
		\definecolor{orchid225122182}{RGB}{225,122,182}
		\definecolor{orchid226139192}{RGB}{226,139,192}
		\definecolor{palevioletred22382154}{RGB}{223,82,154}
		\definecolor{palevioletred225104170}{RGB}{225,104,170}
		\definecolor{plum227155202}{RGB}{227,155,202}
		\definecolor{plum228171210}{RGB}{228,171,210}
		\definecolor{royalblue33118223}{RGB}{33,118,223}
		\definecolor{royalblue46104223}{RGB}{46,104,223}
		\definecolor{royalblue6089216}{RGB}{60,89,216}
		\definecolor{royalblue6876201}{RGB}{68,76,201}
		\definecolor{seagreen24135103}{RGB}{24,135,103}
		\definecolor{seagreen34145102}{RGB}{34,145,102}
		\definecolor{seagreen47155100}{RGB}{47,155,100}
		\definecolor{skyblue117196214}{RGB}{117,196,214}
		\definecolor{skyblue138205215}{RGB}{138,205,215}
		\definecolor{teal18114101}{RGB}{18,114,101}
		\definecolor{teal19125102}{RGB}{19,125,102}
		
		\begin{groupplot}[group style={group size=3 by 2,horizontal sep=0.5cm,vertical sep=0.5cm },height=\textwidth/5,width=\textwidth/3]
		\nextgroupplot[
		scaled x ticks=manual:{}{\pgfmathparse{#1}},
		tick align=outside,
		xtick distance=0.25,
		tick pos=left,
		title={\SI{24}{\hour}},
		x grid style={darkgray176},
		xmin=-0.05, xmax=1.05,
		xtick style={color=black},
		xticklabels={},
		y grid style={darkgray176},
		ylabel={\textit{R2R}},
		ymin=0, ymax=1.5,
		ytick style={color=black}
		]
		\draw[draw=none,fill=darkslategray267386] (axis cs:-6.93889390390723e-18,0) rectangle (axis cs:0.0666666666666667,1.38879946653177);
		\draw[draw=none,fill=darkslategray248391] (axis cs:0.0666666666666667,0) rectangle (axis cs:0.133333333333333,1.00059785233047);
		\draw[draw=none,fill=darkslategray219395] (axis cs:0.133333333333333,0) rectangle (axis cs:0.2,0.95386189611166);
		\draw[draw=none,fill=darkslategray1910399] (axis cs:0.2,0) rectangle (axis cs:0.266666666666667,0.932304720733979);
		\draw[draw=none,fill=teal18114101] (axis cs:0.266666666666667,0) rectangle (axis cs:0.333333333333333,0.912644576789533);
		\draw[draw=none,fill=teal19125102] (axis cs:0.333333333333333,0) rectangle (axis cs:0.4,0.914196693416726);
		\draw[draw=none,fill=seagreen24135103] (axis cs:0.4,0) rectangle (axis cs:0.466666666666667,0.911782289774426);
		\draw[draw=none,fill=seagreen34145102] (axis cs:0.466666666666667,0) rectangle (axis cs:0.533333333333333,0.916266182252984);
		\draw[draw=none,fill=seagreen47155100] (axis cs:0.533333333333333,0) rectangle (axis cs:0.6,0.941617420497137);
		\draw[draw=none,fill=mediumseagreen6416497] (axis cs:0.6,0) rectangle (axis cs:0.666666666666667,0.942479707512245);
		\draw[draw=none,fill=mediumseagreen8817494] (axis cs:0.666666666666667,0) rectangle (axis cs:0.733333333333333,0.960760192232519);
		\draw[draw=none,fill=darkseagreen11418199] (axis cs:0.733333333333333,0) rectangle (axis cs:0.8,0.995251672836808);
		\draw[draw=none,fill=darkseagreen136187112] (axis cs:0.8,0) rectangle (axis cs:0.866666666666667,1.01111775391478);
		\draw[draw=none,fill=darkseagreen154194130] (axis cs:0.866666666666667,0) rectangle (axis cs:0.933333333333333,1.06285497482122);
		\draw[draw=none,fill=darkseagreen170202148] (axis cs:0.933333333333333,0) rectangle (axis cs:1,1.15546460024374);
		\path [draw=black, semithick, dash pattern=on 5.55pt off 2.4pt]
		(axis cs:0,1)
		--(axis cs:1,1);

		\nextgroupplot[
		scaled x ticks=manual:{}{\pgfmathparse{#1}},
		scaled y ticks=manual:{}{\pgfmathparse{#1}},
		tick align=outside,
		xtick distance=0.25,
		tick pos=left,
		title={\SI{72}{\hour}},
		x grid style={darkgray176},
		xmin=-0.05, xmax=1.05,
		xtick style={color=black},
		xticklabels={},
		y grid style={darkgray176},
		ymin=0, ymax=1.5,
		ytick style={color=black},
		yticklabels={}
		]
		\draw[draw=none,fill=darkslateblue6454133] (axis cs:-6.93889390390723e-18,0) rectangle (axis cs:0.0666666661163979,1.37283985818049);
		\draw[draw=none,fill=darkslateblue6860156] (axis cs:0.0666666661163979,0) rectangle (axis cs:0.133333332232796,1.01445276957508);
		\draw[draw=none,fill=darkslateblue7067179] (axis cs:0.133333332232796,0) rectangle (axis cs:0.199999998349194,1.01583250812601);
		\draw[draw=none,fill=royalblue6876201] (axis cs:0.199999998349194,0) rectangle (axis cs:0.266666664465592,0.992376952760115);
		\draw[draw=none,fill=royalblue6089216] (axis cs:0.266666664465592,0) rectangle (axis cs:0.333333330581989,0.997033570369521);
		\draw[draw=none,fill=royalblue46104223] (axis cs:0.333333330581989,0) rectangle (axis cs:0.399999996698387,0.954606609928264);
		\draw[draw=none,fill=royalblue33118223] (axis cs:0.399999996698387,0) rectangle (axis cs:0.466666662814785,0.975647622830026);
		\draw[draw=none,fill=dodgerblue25131222] (axis cs:0.466666662814785,0) rectangle (axis cs:0.533333328931183,1.01807458327128);
		\draw[draw=none,fill=dodgerblue29143220] (axis cs:0.533333328931183,0) rectangle (axis cs:0.599999995047581,0.984098521454503);
		\draw[draw=none,fill=dodgerblue41155217] (axis cs:0.599999995047581,0) rectangle (axis cs:0.666666661163979,1.0091062826902);
		\draw[draw=none,fill=mediumturquoise59167215] (axis cs:0.666666661163979,0) rectangle (axis cs:0.733333327280377,0.97375048232249);
		\draw[draw=none,fill=mediumturquoise77177214] (axis cs:0.733333327280377,0) rectangle (axis cs:0.799999993396775,0.971680874496087);
		\draw[draw=none,fill=mediumturquoise97187213] (axis cs:0.799999993396775,0) rectangle (axis cs:0.866666659513173,0.936497541447242);
		\draw[draw=none,fill=skyblue117196214] (axis cs:0.866666659513173,0) rectangle (axis cs:0.93333332562957,0.892690842455049);
		\draw[draw=none,fill=skyblue138205215] (axis cs:0.93333332562957,0) rectangle (axis cs:0.999999991745968,0.891311103904114);
		\path [draw=black, semithick, dash pattern=on 5.55pt off 2.4pt]
		(axis cs:0,1)
		--(axis cs:1,1);

		\nextgroupplot[
		scaled x ticks=manual:{}{\pgfmathparse{#1}},
		scaled y ticks=manual:{}{\pgfmathparse{#1}},
		tick align=outside,
		xtick distance=0.25,
		tick pos=left,
		title={\SI{120}{\hour}},
		x grid style={darkgray176},
		xmin=-0.05, xmax=1.05,
		xtick style={color=black},
		xticklabels={},
		y grid style={darkgray176},
		ymin=0, ymax=1.5,
		ytick style={color=black},
		yticklabels={}
		]
		\draw[draw=none,fill=maroon1163019] (axis cs:1.82677142857024e-09,0) rectangle (axis cs:0.0666666677850414,1.23441956973541);
		\draw[draw=none,fill=maroon1322824] (axis cs:0.0666666677850414,0) rectangle (axis cs:0.133333333743311,1.02382486599824);
		\draw[draw=none,fill=brown1492431] (axis cs:0.133333333743311,0) rectangle (axis cs:0.199999999701581,0.995193645015133);
		\draw[draw=none,fill=firebrick1661641] (axis cs:0.199999999701581,0) rectangle (axis cs:0.266666665659851,0.961905538691403);
		\draw[draw=none,fill=firebrick181555] (axis cs:0.266666665659851,0) rectangle (axis cs:0.333333331618121,0.941208270510846);
		\draw[draw=none,fill=crimson197175] (axis cs:0.333333331618121,0) rectangle (axis cs:0.399999997576391,0.95621378994175);
		\draw[draw=none,fill=crimson2081996] (axis cs:0.399999997576391,0) rectangle (axis cs:0.466666663534661,0.967769764675894);
		\draw[draw=none,fill=mediumvioletred21541117] (axis cs:0.466666663534661,0) rectangle (axis cs:0.533333329492931,0.97760096706166);
		\draw[draw=none,fill=mediumvioletred22062137] (axis cs:0.533333329492931,0) rectangle (axis cs:0.599999995451201,0.982775284106798);
		\draw[draw=none,fill=palevioletred22382154] (axis cs:0.599999995451201,0) rectangle (axis cs:0.666666661409471,0.987259692212587);
		\draw[draw=none,fill=palevioletred225104170] (axis cs:0.666666661409471,0) rectangle (axis cs:0.73333332736774,1.00347255228736);
		\draw[draw=none,fill=orchid225122182] (axis cs:0.73333332736774,0) rectangle (axis cs:0.79999999332601,1.00899182380217);
		\draw[draw=none,fill=orchid226139192] (axis cs:0.79999999332601,0) rectangle (axis cs:0.86666665928428,1.04779920164072);
		\draw[draw=none,fill=plum227155202] (axis cs:0.86666665928428,0) rectangle (axis cs:0.93333332524255,0.998643189711893);
		\draw[draw=none,fill=plum228171210] (axis cs:0.93333332524255,0) rectangle (axis cs:0.99999999120082,0.912922003997417);
		\path [draw=black, semithick, dash pattern=on 5.55pt off 2.4pt]
		(axis cs:0,1)
		--(axis cs:1,1);

		\nextgroupplot[
		tick align=outside,
		xtick distance=0.25,
		tick pos=left,
		x grid style={darkgray176},
		xlabel={PIT value},
		xmin=-0.05, xmax=1.05,
		xtick style={color=black},
		y grid style={darkgray176},
		ylabel={\textit{R2F}},
		ymin=0, ymax=1.5,
		ytick style={color=black}
		]
		\draw[draw=none,fill=darkslategray267386] (axis cs:-6.93889390390723e-18,0) rectangle (axis cs:0.0666666550928434,1.2414403483308);
		\draw[draw=none,fill=darkslategray248391] (axis cs:0.0666666550928434,0) rectangle (axis cs:0.133333310185687,1.00487670485448);
		\draw[draw=none,fill=darkslategray219395] (axis cs:0.133333310185687,0) rectangle (axis cs:0.19999996527853,0.969253817021056);
		\draw[draw=none,fill=darkslategray1910399] (axis cs:0.19999996527853,0) rectangle (axis cs:0.266666620371374,0.97340405638029);
		\draw[draw=none,fill=teal18114101] (axis cs:0.266666620371373,0) rectangle (axis cs:0.333333275464217,0.968735037101152);
		\draw[draw=none,fill=teal19125102] (axis cs:0.333333275464217,0) rectangle (axis cs:0.39999993055706,0.959742851822811);
		\draw[draw=none,fill=seagreen24135103] (axis cs:0.39999993055706,0) rectangle (axis cs:0.466666585649904,0.978591855579332);
		\draw[draw=none,fill=seagreen34145102] (axis cs:0.466666585649904,0) rectangle (axis cs:0.533333240742747,1.00591426469429);
		\draw[draw=none,fill=seagreen47155100] (axis cs:0.533333240742747,0) rectangle (axis cs:0.59999989583559,0.986892334297799);
		\draw[draw=none,fill=mediumseagreen6416497] (axis cs:0.599999895835591,0) rectangle (axis cs:0.666666550928434,1.01525230325256);
		\draw[draw=none,fill=mediumseagreen8817494] (axis cs:0.666666550928434,0) rectangle (axis cs:0.733333206021277,1.01283133029301);
		\draw[draw=none,fill=darkseagreen11418199] (axis cs:0.733333206021277,0) rectangle (axis cs:0.799999861114121,0.996576226136011);
		\draw[draw=none,fill=darkseagreen136187112] (axis cs:0.799999861114121,0) rectangle (axis cs:0.866666516206964,0.985854774457992);
		\draw[draw=none,fill=darkseagreen154194130] (axis cs:0.866666516206964,0) rectangle (axis cs:0.933333171299807,0.988794527337448);
		\draw[draw=none,fill=darkseagreen170202148] (axis cs:0.933333171299807,0) rectangle (axis cs:0.999999826392651,0.911842172551655);
		\path [draw=black, semithick, dash pattern=on 5.55pt off 2.4pt]
		(axis cs:0,1)
		--(axis cs:1,1);

		\nextgroupplot[
		scaled y ticks=manual:{}{\pgfmathparse{#1}},
		tick align=outside,
		xtick distance=0.25,
		tick pos=left,
		x grid style={darkgray176},
		xlabel={PIT value},
		xmin=-0.05, xmax=1.05,
		xtick style={color=black},
		y grid style={darkgray176},
		ymin=0, ymax=1.5,
		ytick style={color=black},
		yticklabels={}
		]
		\draw[draw=none,fill=darkslateblue6454133] (axis cs:6.93889390390723e-18,0) rectangle (axis cs:0.0666652124525034,1.10139360368344);
		\draw[draw=none,fill=darkslateblue6860156] (axis cs:0.0666652124525034,0) rectangle (axis cs:0.133330424905007,0.982590431171194);
		\draw[draw=none,fill=darkslateblue7067179] (axis cs:0.133330424905007,0) rectangle (axis cs:0.19999563735751,1.03187559007366);
		\draw[draw=none,fill=royalblue6876201] (axis cs:0.19999563735751,0) rectangle (axis cs:0.266660849810014,1.02288321020373);
		\draw[draw=none,fill=royalblue6089216] (axis cs:0.266660849810014,0) rectangle (axis cs:0.333326062262517,1.080987818594);
		\draw[draw=none,fill=royalblue46104223] (axis cs:0.333326062262517,0) rectangle (axis cs:0.399991274715021,1.05573998280538);
		\draw[draw=none,fill=royalblue33118223] (axis cs:0.399991274715021,0) rectangle (axis cs:0.466656487167524,1.07078492604928);
		\draw[draw=none,fill=dodgerblue25131222] (axis cs:0.466656487167524,0) rectangle (axis cs:0.533321699620028,1.0759728375127);
		\draw[draw=none,fill=dodgerblue29143220] (axis cs:0.533321699620027,0) rectangle (axis cs:0.599986912072531,1.09326587572409);
		\draw[draw=none,fill=dodgerblue41155217] (axis cs:0.599986912072531,0) rectangle (axis cs:0.666652124525034,1.05746928662651);
		\draw[draw=none,fill=mediumturquoise59167215] (axis cs:0.666652124525034,0) rectangle (axis cs:0.733317336977538,1.0249583747891);
		\draw[draw=none,fill=mediumturquoise77177214] (axis cs:0.733317336977538,0) rectangle (axis cs:0.799982549430041,0.997289513650877);
		\draw[draw=none,fill=mediumturquoise97187213] (axis cs:0.799982549430041,0) rectangle (axis cs:0.866647761882545,0.923621170870353);
		\draw[draw=none,fill=skyblue117196214] (axis cs:0.866647761882545,0) rectangle (axis cs:0.933312974335048,0.833870302553237);
		\draw[draw=none,fill=skyblue138205215] (axis cs:0.933312974335048,0) rectangle (axis cs:0.999978186787552,0.647624281016564);
		\path [draw=black, semithick, dash pattern=on 5.55pt off 2.4pt]
		(axis cs:0,1)
		--(axis cs:1,1);

		\nextgroupplot[
		scaled y ticks=manual:{}{\pgfmathparse{#1}},
		tick align=outside,
		xtick distance=0.25,
		tick pos=left,
		x grid style={darkgray176},
		xlabel={PIT value},
		xmin=-0.05, xmax=1.05,
		xtick style={color=black},
		y grid style={darkgray176},
		ymin=0, ymax=1.5,
		ytick style={color=black},
		yticklabels={}
		]
		\draw[draw=none,fill=maroon1163019] (axis cs:3.33966319876478e-08,0) rectangle (axis cs:0.0666666441548684,0.960608123845517);
		\draw[draw=none,fill=maroon1322824] (axis cs:0.0666666441548684,0) rectangle (axis cs:0.133333254913105,0.952480566362035);
		\draw[draw=none,fill=brown1492431] (axis cs:0.133333254913105,0) rectangle (axis cs:0.199999865671341,1.00591493364705);
		\draw[draw=none,fill=firebrick1661641] (axis cs:0.199999865671341,0) rectangle (axis cs:0.266666476429578,1.00799005470666);
		\draw[draw=none,fill=firebrick181555] (axis cs:0.266666476429578,0) rectangle (axis cs:0.333333087187814,1.01923029377956);
		\draw[draw=none,fill=crimson197175] (axis cs:0.333333087187814,0) rectangle (axis cs:0.399999697946051,1.03358321444188);
		\draw[draw=none,fill=crimson2081996] (axis cs:0.399999697946051,0) rectangle (axis cs:0.466666308704287,1.03531248199156);
		\draw[draw=none,fill=mediumvioletred21541117] (axis cs:0.466666308704287,0) rectangle (axis cs:0.533332919462524,1.03548540874653);
		\draw[draw=none,fill=mediumvioletred22062137] (axis cs:0.533332919462524,0) rectangle (axis cs:0.59999953022076,1.02528273020343);
		\draw[draw=none,fill=palevioletred22382154] (axis cs:0.59999953022076,0) rectangle (axis cs:0.666666140978997,1.04344003947504);
		\draw[draw=none,fill=palevioletred225104170] (axis cs:0.666666140978997,0) rectangle (axis cs:0.733332751737233,1.05554491232278);
		\draw[draw=none,fill=orchid225122182] (axis cs:0.733332751737233,0) rectangle (axis cs:0.799999362495469,1.06868734570032);
		\draw[draw=none,fill=orchid226139192] (axis cs:0.79999936249547,0) rectangle (axis cs:0.866665973253706,1.04914662238897);
		\draw[draw=none,fill=plum227155202] (axis cs:0.866665973253706,0) rectangle (axis cs:0.933332584011943,0.980148847156864);
		\draw[draw=none,fill=plum228171210] (axis cs:0.933332584011942,0) rectangle (axis cs:0.999999194770179,0.727157004639136);
		\path [draw=black, semithick, dash pattern=on 5.55pt off 2.4pt]
		(axis cs:0,1)
		--(axis cs:1,1);
		
		\end{groupplot}
		
		\end{tikzpicture}
		\caption{PIT histograms of the post-processed forecasts of the DRN and GAT model for \SIlist{24;72;120}{\hour} lead times based on the \textit{R2R} and \textit{R2F} tasks.}
		\phantomsection
		\label{fig:PIT}
	\end{center}
	\vskip -0.2in
\end{figure}

To assess the calibration of the different post-processing approaches, we use probability integral transform (PIT) histograms.
The PIT $F(y)$ is the value of the predictive CDF $F$, evaluated at the \feat{t2m} observation  $y$. 
In our case, the predictive distribution is Gaussian and thus 
$
F \left( y \right) = \operatorname{\Phi}\left( \frac{y-\mu}{\sigma}\right)
$
is the PIT.
If the model is calibrated, meaning the realizing \feat{t2m} observation is indistinguishable from a random draw from the forecast distribution, the PIT values should follow a uniform distribution $\mathcal{U}\left(0,1\right)$, and the visual inspection of histograms of the PIT values can point to different kinds of mis-calibration. For example, histograms that follow a U-shape indicate that the forecast is underdispersive (i.e., the observation too often falls outside a plausible predicted range). 
\Cref{fig:PIT} shows PIT histograms of the DRN and GAT models for the different lead times. 
All PIT histograms resemble an uniform distribution fairly well, however, for the reforecast data there exists a spike for the lower PIT values and larger PIT values are under-presented, specifically for the \textit{R2F} task. 
Overall, only minor differences between the PIT histograms of the DRN and the GAT model can be observed.

\subsubsection{Feature Importance}
\label{app:feature_imp}
To identify the most important input features, we employ a permutation importance approach, which operates on the fundamental assumption that an input feature's importance can be determined by measuring the impact of randomly shuffling it on the model's performance.
If an input variable is important, the predictive performance deteriorates notably after permuting it, while for unimportant variables, performance remains relatively unchanged.
This can be due to the variable being generally unimportant for the task at hand, or the redundancy of the variable, meaning the information of this variable is already captured by other variables through multicollinearities \citep{mcgovern_making_2019}.
The main advantage is that the model does not have to be retrained each time, saving computational resources. However, colinearities or interactions between variables cannot be captured.

Following \citet{hohlein_postprocessing_2024}, we employ a two-step permutation is employed to first permute the feature across the time dimension and subsequently across the station ($s$) and ensemble member ($n$) dimension to evaluate the importance of an input variable $i$. Let
\begin{equation*}
\mathbf{X}_t = \{\mathbf{x}_{t,s,n} \vert s=1,\dots,S;\,n=1,\dots,N\}
\end{equation*}
denote the entire dataset at time step $t$, where $\mathbf{x}_{t,s,n}$ is a  vector in $\mathbb{R}^P$ describing the prediction at station $s$ and time $t$, made by ensemble member $n$, and $P$ is the total number of input features.
To simplify notation, we omit the index for the $i$-th feature, however note that the following transformations are only applied to the $i$-th dimension of $\mathbf{x}_{t,s,n}$. 
First, the data is permuted along the time dimension, according to a permutation $\pi$. 
Second, for each time-stamp $t$, the feature of interest is permuted along the station and ensemble member dimension together.
Therefore,
\begin{equation}
\Pi(\mathbf{X}_{t}) = \{\mathbf{x}_{\pi(t),\pi_t(s,n)}\vert s=1,\dots,S;\,n=1,\dots,N\}
\end{equation}
is the permuted feature set, which is then used to generate the graphs, as detailed in \Cref{sec:graph topology}.
The two-stage shuffling is designed to maintain certain structural information in the graph, such as ensuring that each station ID appears an equal number of times each day, irrespective of the shuffling.
The importance of each feature is calculated by comparing the mean CRPS of the permuted dataset and to the original one computed on the non-permuted data via
\begin{equation}
\label{eq:imp}
\operatorname{Imp}(i) = \frac{\overline{\operatorname{CRPS}}(\mathbf{F}\vert\Pi_i(\mathbf{X}), \mathbf{Y})-\overline{\operatorname{CRPS}}(\mathbf{F}\vert\mathbf{X}, \mathbf{Y})}{\overline{\operatorname{CRPS}}(\mathbf{F}\vert\mathbf{X}, \mathbf{Y})}.
\end{equation}
The importance of feature $i$ is estimated by evaluating \Cref{eq:imp} 10 times using a different training run of a single GNN model. 

\begin{figure}[ht]
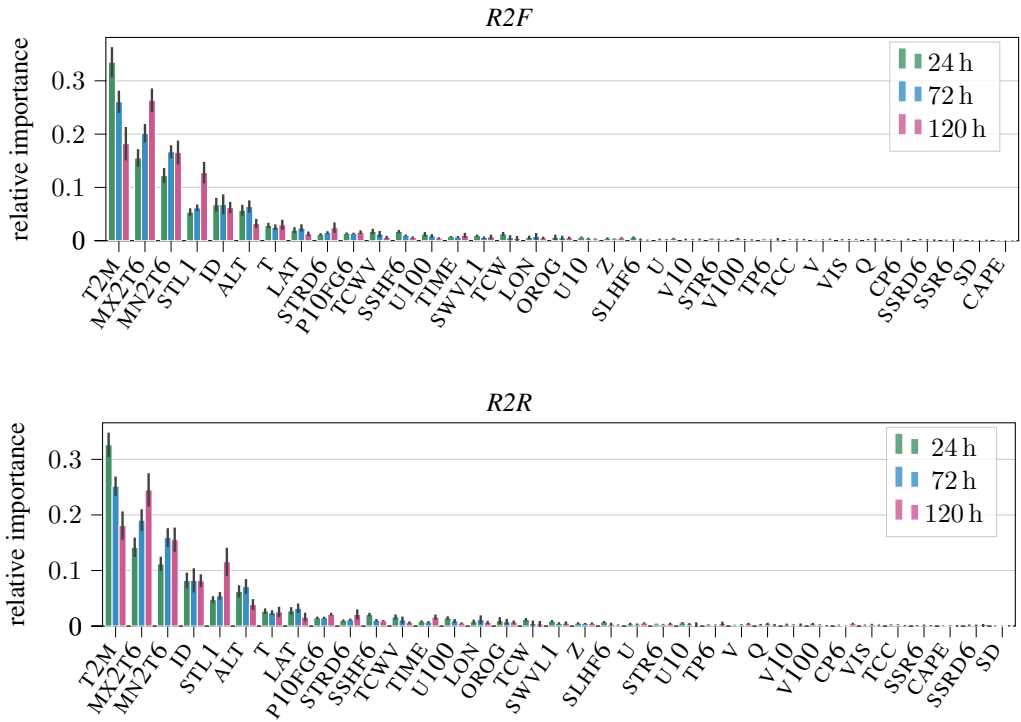

	\vskip 0.2in
	\begin{center}
		\textit{R2F} \\

		\caption{Relative feature importance of the GAT model for the \textit{R2F} (top) and the \textit{R2R} task (bottom). Error bars show the standard deviation, which is calculated based on 10 training runs of the individual GNNs.}
		\phantomsection
		\label{fig:PI_f_full}
	\end{center}
	\vskip -0.2in
\end{figure}

\Cref{fig:PI_f_full} shows the feature importance for the \SIlist{24;72;120}{\hour} lead times for the two tasks. Note that the feature importances are  normalized to allow for a better comparison.
Not surprisingly, the top 3 most important predictor variables (\sc t2m, mx2t6, mn2t6\normalfont) all concern the 2-m temperature and account for about \SI{61.8}{\percent} of the total importance together (in the \textit{R2F} task). 
Even though the distribution of importances across these three variables varies substantially depending on the lead time, the total importance always sums up to \SI{61.8}{}$\pm$\SI{1}{\percent}. 
For the \textit{R2F} task, the temperature variables are followed by the level 1 soil temperature (\sc stl1\normalfont), which is recorded in a a depth of \SIrange{0}{7}{\centi\meter}.
As the lead time increases, the importance of soil temperature increases as well. 
Subsequently, two station-specific features follow, where \sc id \normalfont refers to the station identifier, which is arbitrarily assigned in the beginning, but is mapped via the embedding layer to a 20-dimensional vector. Using this embedding, the model encodes station specific information in the node id during training. \sc alt \normalfont refers to the altitude of the station location.
Qualitatively similar results are obtained for the \textit{R2R} task, with a change in the raking between the soil temperature and the station identifier being the most notable difference in the most important predictors.
However, note that these feature importances should be interpreted with care, as the quality of the prediction made by the NWP model varies across the features. Thus low importance can also be due to decreased forecasting performance by the NWP model, instead of the variable being irrelevant for the task. For details, see also the corresponding discussions in \citet{rasp_neural_2018} and \citet{schulz_machine_2022}.

\end{document}